# Deep Q-learning of global optimizer of multiply model parameters from nonconvex function $f$

Hongmei Zhang, Kai Wang, Yan Zhou, Shadab Momin, Xiaofeng Yang, Mostafa Fatemi, Michael F. Insana

*Abstract—Objective:* Estimation of the global optima of multiple model parameters is valuable in imaging to form a reliable diagnostic image. Given non convexity of the objective function, it is challenging to avoid from different local minima. *Methods:* We first formulate the global searching of multiply parameters to be a k-D move in the parametric space, and convert parameters updating to be state-action decision-making problem. We proposed a novel Deep Q-learning of Model Parameters (DQMP) method for global optimization of model parameters by updating the parameter configurations through actions that maximize a Q-value, which employs a Deep Reward Network designed to learn global reward values from both visible curve fitting errors and hidden parameter errors. *Results:* The DQMP method was evaluated by viscoelastic imaging on soft matter by Kelvin-Voigt fractional derivative (KVFD) modeling. In comparison to other methods, imaging of parameters by DQMP yielded the smallest errors (< 2%) to the ground truth images. DQMP was applied to viscoelastic imaging on biological tissues, which indicated a great potential of imaging on physical parameters in diagnostic applications. *Conclusions:* DQMP method is able to achieve global optima, yielding accurate model parameter estimates in viscoelastic imaging. Assessment of DQMP by simulation imaging and ultrasound breast imaging demonstrated the consistency, reliability of the imaged parameters, and powerful global searching ability of DQMP. *Significance:* DQMP method is promising for imaging of multiple parameters, and can be generalized to global optimization for many other complex nonconvex functions and imaging of physical parameters.

*Index Terms—*Deep Q-learning; Q-value; Viscoelastic imaging; Parameters optimization; Global optima.

## I. INTRODUCTION

MODEL fitting is a branch of nonlinear regression problems that simultaneously extracts multiple model parameters by fitting experiment data to a specific model. Estimating multiple model parameters from response curve are of great importance in many measurement and imaging applications [1]-[3]. When the fitting function $f$ is nonconvex and there are few known constraints, achieving global convergence for parameter estimation is challenging.

There is no general algorithm for solving these problems, and the theoretical guarantees regarding convergence to global optima for common algorithms are weak or nonexistent. The established field of optimization is extensive, consisting of basic methods such as Gauss-Newton and gradient descent [4], [5], as well as combinations of those methods, such as Levenberg-Marquardt (LM) [6], [7]. Each of these methods have their strengths and weaknesses.

The simulated annealing algorithm is a stochastic scheme for searching global optimum by minimizing the system of energy through an annealing schedule [8]. Genetic algorithm is bio-inspired heuristic search through heredity and mutation [9], [10]. Particle swarm is another bio-inspired stochastic approach based on the best positions experienced so far by each particle and the whole swarm [11].

Current stochastic and evolutionary schemes can, in theory, jump out of local extrema and increase the probability of finding global solutions. However, jumping from current local extrema may introducing other local ones, ultimately yielding inconsistent solutions unless it can be guided by any valid prior knowledge that might be available and by relevant past experiences.

Deep learning (DL) methods have the capacity to learn from prior knowledge. Deep neural networks can establish maps from input to output and they may be scaled to model arbitrary mappings. The adaptive and nonlinear response of their neural networks can be trained to model highly complex systems. With the successful application of DL in AlphaGo [12], [13], the power of DL has been validated in a variety of applications [14]-[20].

In recent years, DL applied to regression task was reportedly capable of solving multi-parameter optimization and curve fitting problems [19]-[22]. However, learning a large number of model parameters and network weights is a complex optimization problem itself due to its network-like nature.

This study was financially supported by National Natural Science Foundation of China (62171366, 61871316) and the Fundamental Research Funds for the Central Universities (zdyf2017011). *(Corresponding authors: Hongmei Zhang and Michael F. Insana.)*

Hongmei Zhang is with Key Laboratory of Biomedical Information Engineering of Ministry of Education, School of Life Science and Technology, Xi'an Jiaotong University, Xi'an, Shaanxi 710049, China (e-mail: claramei@mail.xjtu.edu.cn).

Kai Wang is with School of Cyber Science and Engineering, Xi'an Jiaotong University, China.

Yan Zhou is with Key Laboratory of Biomedical Information Engineering of Ministry of Education, School of Life Science and Technology, Xi'an Jiaotong University, China.

Shadab Momin and Xiaofeng Yang are with Emory University Winship Cancer Institute, Department of Radiation Oncology, Emory University, USA.

Mostafa Fatemi is with Mayo Clinic Department of Physiology and Biomedical Engineering, Mayo Clinic, USA.

Michael F. Insana is with Beckman Institute for Advanced Science and Technology, Department of Bioengineering, University of Illinois at Urbana-Champaign, Urbana, IL 61801, USA (e-mail: mfi@illinois.edu).

Convergence may be difficult to achieve when applying DL to learn multiple model parameters [19], [22].

Feedback from past activity is important to human learning. Reinforcement Learning (RL) is an agent-based powerful AI algorithm in which the agents learn the optimal set of actions through their interaction with the environment. RL is able to make appropriate responses because of reinforcing events. These events can include human feedback to responses that includes rewards and punishments as quantified by a value function. In this way, past experiences can guide RL in learning similar new experiences. RL is a promising branch of AI. The essence of RL is to take actions that maximize the value function. [12], [13], [23]-[27].

Q-learning [28] is a model free agent-based RL method that can adapt to environment, including prior knowledge gleaned from past experiences. The key idea of Q-Learning is its value function. The algorithm seeks to be rewarded and to avoid punishment for its current and next action in the form of an increasing value function. Due to a cumulative feedback mechanism [28], the agent learns to associate the optimal action for each state [29] in pursuit of increasing its value function. Q-learning is widely used in decision-making, gambling, and random event processing problems [30]-[32]. Combinations of DL and RL/Q-learning have been successfully implemented to solve complex human activities, for example, AlphaGo [12], [13]. A deep Q-Network can learn from prior human experience and predict the value function through training.

The aim of this paper is to develop a novel Deep Q-learning of Model Parameters (DQMP) algorithm that finds a global optima for estimating multiply model parameters when the objective function is nonconvex. We first convert the search for global parameters into a decision-making problem in parameters space based on both visible (curve fitting) and hidden (parameters fitting) states. For a $k$-parameter optimization problem, we subdivide the next action into $2^k$ directions in the parameters space. Actions involve selecting a next candidate move in parameter space based on the state of the current model fit, including both the visible state and the hidden state feedbacks. In this way, we formulate model parameter optimization as an action selection problem in parameter space.

To combine data with prior knowledge, a Deep Reward Network (DRN) is proposed to learn the global reward function. This process integrates both visible and hidden state feedbacks. Then a novel DQMP scheme is proposed to maximize the Q-value function. This strategy guides the DQMP search towards the global solution.

DQMP is applied to viscoelastic measurement from soft tissues with the goal of estimating Kelvin-Voigt fractional derivative (KVFD) model parameters. The assessment and validation of DQMP is tested using simulated data, nanoindentation data, and in vivo ultrasonic data with KVFD modeling.

## II. METHODS

Let $\boldsymbol{\theta} = [\theta_1, \theta_2, \ldots, \theta_k]$ be a $k$-dimensional parameter vector for a time-dependent material model that predicts the measurements, $\boldsymbol{Y} = f(t; \boldsymbol{\theta})$. If $\widetilde{\boldsymbol{Y}}$ is a vector of measurement data from an experiment, then $\boldsymbol{\theta}$ represents properties of the material if $\widetilde{\boldsymbol{Y}} \approx \boldsymbol{Y} = f(t; \boldsymbol{\theta})$. Our ultimate goal is to estimate $\boldsymbol{\theta}$.

Let $\widehat{\boldsymbol{\theta}}$ be an estimate of $\boldsymbol{\theta}$, and $\widehat{\boldsymbol{Y}} = f(t; \widehat{\boldsymbol{\theta}})$ is a model of the measurement data that is based on that estimate. Let $\beta_1$ and $\beta_2$ be weights that balance the contributions of two fitting error terms describing model parameters and measurement curves, respectively. The global optimum is estimated by minimizing the following objective function:

$$\widehat{\boldsymbol{\theta}} = \underset{\widehat{\boldsymbol{\theta}}}{\mathrm{argmin}} \left( \beta_1 \underbrace{|\boldsymbol{\theta} - \widehat{\boldsymbol{\theta}}|}_{hidden} + \beta_2 \underbrace{|f(t; \boldsymbol{\theta}) - f(t; \widehat{\boldsymbol{\theta}})|}_{visible} \right). \quad (1)$$

In (1), the *visible* term involves measurable data that depends on the parameters, whereas the *hidden* term evaluates the parameters.

We convert the parameter optimizing to a decision-making problems elegantly by minimize (1) through a set of state-action decisions in parameter space. In the fitting problem, the state are referred to $s: \{\widehat{\boldsymbol{\theta}}, f(t; \widehat{\boldsymbol{\theta}})\}$ including (1) visible state $f(t; \widehat{\boldsymbol{\theta}})$ on which the difference between current curve and the desired curve is measured by $|f(t; \boldsymbol{\theta}) - f(t; \widehat{\boldsymbol{\theta}})|$; 2) hidden state $\widehat{\boldsymbol{\theta}}$ on which the errors between the current parameter configuration $\widehat{\boldsymbol{\theta}}$ and the true parameter configuration $\boldsymbol{\theta}$ is measured by $|\boldsymbol{\theta} - \widehat{\boldsymbol{\theta}}|$.

There are $n = 2^k$ candidate actions possible at the current state in $k$-D parameter space. To help understand the problem setting, take $k=3$ for example. As listed in Fig. 1, the current parameter state (yellow dot) $\widehat{\boldsymbol{\theta}}$ can move in one of eight possible directions (red dots) $\widehat{\boldsymbol{\theta}}_{a_1}, \ldots, \widehat{\boldsymbol{\theta}}_{a_8}$ in the next step by taking actions $a_1, \ldots, a_8$ respectively. In this way, updating of $\widehat{\boldsymbol{\theta}}$ in the parametric space resembles selection move in the 3D board for game, which is a novel approach for optimizing model parameters.

The optimal actions may be taken to increase a value function as $\widehat{\boldsymbol{\theta}} \rightarrow \boldsymbol{\theta}$. Q-value is the expected discounted reward for executing action $a$ at state $s$ and the next step optimal action $a'$ at state $s'$ by episodes thereafter. Q-Learning is a powerful scheme for agents to learn to act optimally by experiencing the consequences of actions judged by long-term discounted reward, in which selecting every action can obtain the maximum benefits-Q value.

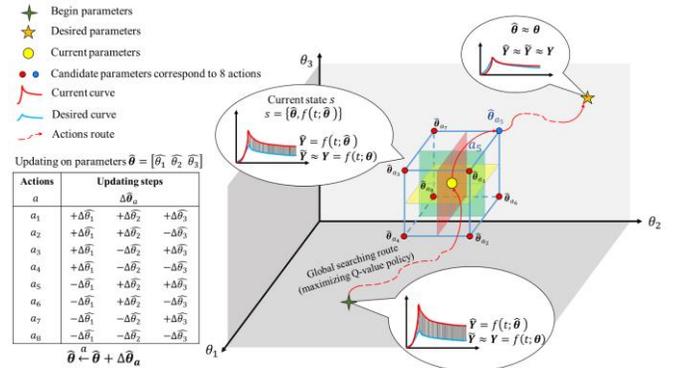

Fig. 1. Schematic illustration of a global search of state-actions in a 3-D parameter space. The yellow center point denotes the current parameter configuration $\widehat{\boldsymbol{\theta}}$, and the 8 edge points are the candidate actions. The decision to select an action is based on maximizing the Q-value policy. The orange star point denotes the next move through action $a_5$ that may maximize the Q-value function.

Q-learning consists of a set of states $S$, a set of actions $A$, and a reward function $r: S \times A \to R^+$. It uses the Q-value to feedback the actions of every step. The policy $\pi$ maps states to actions as $\pi: S \to A$. The policy is maximizing Q-value by [28]

$$Q(s,a) \leftarrow \underbrace{Q(s,a)}_{cumulative} + \xi \left[ \underbrace{r(s,a)}_{immediate} + \gamma \underbrace{\max_{a'} Q(s',a')}_{future} - Q(s,a) \right], \quad (2)$$

$r(s,a)$ is the immediate reward of selecting action $a$ at distinct state $s$. The reward can be any positive value for rewarding an action. The reward function should be a decreasing function of the fitting errors in the fitting problems. $Q(s',a')$ is the $Q$ value found by selecting the next state-action pairs $(s',a')$. $\xi \in [0,1]$ is the learning rate and $\gamma \in [0,1]$ is a discount factor. In this work, $\xi$ is 0.6 and $\gamma$ is 0.5.

In this way, we formulate the parameter search to be a state-action decision-making in the parameter space by increasing the Q-value function policy. The parameter update is formulated to be $a = \underset{a_i}{\operatorname{argmax}} Q(s, a_i)$ at every step, then parameters are updated by $\widehat{\boldsymbol{\theta}} \xleftarrow{a} \widehat{\boldsymbol{\theta}} + \Delta \widehat{\boldsymbol{\theta}_a}$ as indicated in the left table of Fig. 1. The adaptive steps for updating parameter are the 1% of the current parameter.

To guide the global parameter search, the Q-value function should integrate both the curve fitting (visible) and parameter fitting (hidden) feedbacks.

In order to learn the prior rewards from hidden states, a Deep Reward Network (DRN) is proposed to learn the reward function whose global constraints are absorbed from both visible and hidden states. In this way, a novel DQMP algorithm is proposed, where a deep initial network (DIN) applies experimental data to initialize $\widehat{\boldsymbol{\theta}}$, while a DRN is to predict global reward value consisting of both the curve fitting (visible reward) and parameters fitting (hidden reward) feedbacks. DQMP iteratively updates the state through convergence such that a global solution can be found following the maximizing Q-value policy.

*A. Deep Q-learning of Model Parameters (DQMP) from nonconvex function*

*1) Q-value integrating rewards from both hidden and visible states*

Q-value is inherently weighted sum of the immediate reward, cumulative reward and the future reward. Learning of reward function $r(s,a)$ that rewards both visible and hidden state feedbacks is crucial to guide the global search. Let $R_{curve}$ denote the curve fitting reward (visible) and $R_\theta$ the parameters fitting reward (hidden). The reward function $r(s,a)$ is formulated as (3), where the global reward $R_g$ combines both the hidden reward $R_\theta$ and visible reward $R_{curve}$ as expressed by (4).

$$r(s,a) = \beta_g R_g + (1-\beta_g) R_{curve} \quad (3)$$

$$R_g = \beta_\theta R_\theta + \beta_c R_{curve} \quad (4\text{-}1)$$

$$R_\theta = 1 - \frac{\|\widehat{\boldsymbol{\theta}} - \boldsymbol{\theta}\|_2}{\sqrt{k}} \quad (4\text{-}2)$$

$$R_{curve} = \begin{cases} 0, & \overline{|\Delta|} > e_{max} \\ g(\overline{|\Delta|}), & e_{min} \le \overline{|\Delta|} \le e_{max} \\ 1, & \overline{|\Delta|} < e_{min} \end{cases} \quad (4\text{-}3)$$

$\beta_g \in [0,1]$ is the weight to balance the global reward and the curve fitting reward. $\beta_\theta, \beta_c \in [0,1]$ are adjustable weights for $R_\theta$ and $R_{curve}$, respectively. $k$ is the number of parameters. $g(\cdot)$ is a decreasing function, $e_{min}$ and $e_{max}$ are the two thresholds such that $g(e_{min}) = 1$ and $g(e_{max}) = 0$, and $\overline{|\Delta|} = \mathbf{E}(|\widehat{Y} - \widetilde{Y}|)$ is expectation of Mean Absolute Error (MAE) between the current curve and desired curve. In this work, $\beta_g$ is 0.02, $\beta_\theta$ is 0.6, $\beta_c$ is 0.4, $e_{min}$ is $10^{-10}$ and $e_{max}$ is 1.

Applying (3) and (4) to (2), a novel Deep Q-learning method is then proposed, which updates the Q-value as expressed by

$$Q(s,a) \leftarrow Q(s,a) + \xi \Big[ \underbrace{\beta_g \underbrace{(\beta_\theta R_\theta + \beta_c R_{curve})}_{R_g} + (1-\beta_g) R_{curve}}_{r(s,a)} + \gamma \max_{a'} Q(s',a') - Q(s,a) \Big]. \quad (5)$$

In the immediate fitting environment, $R_g$ is the global rewards that can be learned and predicted by the DRN.

*2) Deep Reward Network*

The Deep Reward Network (DRN) was designed to predict $R_g$ that consists of reward from both visible and hidden states. The schematic illustration of Deep Reward Network is shown in the Fig. 2.

The generation of training data set is displayed on the left of Fig. 2. First, two parameter configurations, true parameter $\boldsymbol{\theta}$ and current parameter $\widehat{\boldsymbol{\theta}}$, were randomly chosen in the parameter space. The two curves, generated by $\boldsymbol{\theta}$ and $\widehat{\boldsymbol{\theta}}$ according to function $f$, exactly simulates the desired experimental curve $f(t; \boldsymbol{\theta})$ and the immediate curve $f(t; \widehat{\boldsymbol{\theta}})$, respectively. The current parameters $\widehat{\boldsymbol{\theta}}$ perform actions $a_i (i = 1, \ldots, 8)$ to achieve 8 candidate parameter configurations $\hat{\theta}_{a_i}(i = 1, ..., 8)$. With that, the corresponding immediate curve is generated by $f(t; \hat{\theta}_{a_i})$. For each candidate action $a_i$, using (4-2) and (4-3), we can calculate the $R_{curve}$ and $R_\theta$, and hence $R_g$. In this way, how to reward the curve fitting by doing actions $a_i(i = 1, ..., 8)$ is recorded in global rewards $R_g$ as illustrated on the left of Fig. 2.

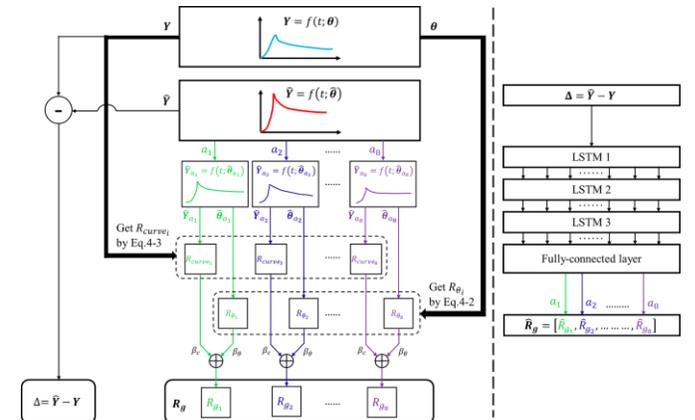

Fig. 2. Schematic illustration of Deep Reward Network (DRN) to learn and predict global reward. The left figure shows the flow of generating training data, and the right figure shows the structure of DRN, which has three LSTM layers followed by a fully-connected layer. The last hidden layer applies a sigmoid nonlinearity, and other hidden layers are followed by a rectifier nonlinearity.

The structure of DRN is shown on the right of Fig. 2. Long Short-Term Memory (LSTM) neural network is used to construct the DRN. LSTM is a special Recurrent Neural Network (RNN) that is appropriate at dealing with sequence data modeling. Comparing with ordinary RNN, LSTM architecture is good at dealing with long time sequence data, unlimited state numbers, and avoid problems related to vanishing and exploding gradients [33]. The input of the Deep Reward Network (DRN) is the difference between the current curve and desired curve ($\Delta(s) = \hat{Y} - \tilde{Y}$), followed by three LSTM layers and a fully-connected layer. The output of the DRN is the global rewards $\{\hat{R}_{g_1}, ..., \hat{R}_{g_n}\}$ ($n = 2^k$) for each action. The last hidden layer is followed by a sigmoid nonlinearity, and the other hidden layers are followed by a rectifier nonlinearity.

The loss function $L_{Reward}$ of Deep Reward Network is given by

$$L_{Reward}(R_g, \hat{R}_g) = \sum_{i=1}^{n} |R_{g_i} - \hat{R}_{g_i}|. \qquad (6)$$

Totally 1,000,000 pair-wise parameters $\theta$ and $\hat{\theta}$ are generated in the training set. We randomly split the datasets into training, validation and test sets with proportions 80%, 10% and 10%, respectively. After training, DRN expects to predict global reward $R_g$ when provided with the current curve and the desired experimental curve in every immediate fitting steps.

*3) Deep Initial Network*

To speed up the convergence, initial guess of parameters is learned from a Deep Initial Network (DIN). The schematic illustration of DIN is shown in Fig. 3. The input of the DIN is curve $\tilde{Y}$, followed by two LSTM layers and a fully-connected layer. The output of the network is the parameters $\hat{\theta}$ which roughly approach global solution $\theta$. The last hidden layer is followed by a sigmoid nonlinearity and other hidden layers are followed by a rectifier nonlinearity. Totally 1,000,000 training data are generated to train DIN.

The loss function integrates the prior knowledge from both curve fitting and parameters fitting are as follows.

$$\begin{cases} L_{Initial}(\theta, Y, \hat{\theta}, \hat{Y}) = \beta_{L\theta} L_\theta + \beta_{Lc} L_c \\ L_\theta(\theta, \hat{\theta}) = \sum_{i=1}^{k} \beta_i \frac{|\hat{\theta}_i - \theta_i|}{\theta_i} = \langle \beta, |\hat{\theta} - \theta| \rangle \\ L_c(Y, \hat{Y}) = \frac{E(|\hat{Y}-Y|)}{\max(Y)} = \frac{E(|f(\hat{\theta})-f(\theta)|)}{\max(f(\theta))} \end{cases}, \qquad (7)$$

where $L_\theta$ and $L_c$ are loss functions from the parameters error and curve error respectively, $\beta_i$ is weight for the *i-th* parameter, and $\hat{\theta}_i$ is the *i-th* parameter to be determined, $\beta_{L\theta}$ and $\beta_{Lc}$ are weight for $L_\theta$ and $L_c$ respectively. In this work, $\beta_{L\theta}$ and $\beta_{Lc}$ are both 1, $\beta_i$ for $E_0$, $\alpha$, and $\tau$ are 0.09, 0.9 and 0.01 in DIN.

*4) DQMP algorithm*

Schematic illustration of DQMP is shown in Fig. 4, where a DIN is used to initialize the parameter vector $\theta$, and a DRN is used to predict the global rewards function, $R_g$. The pseudo-code of DQMP is provided in Appendix A.

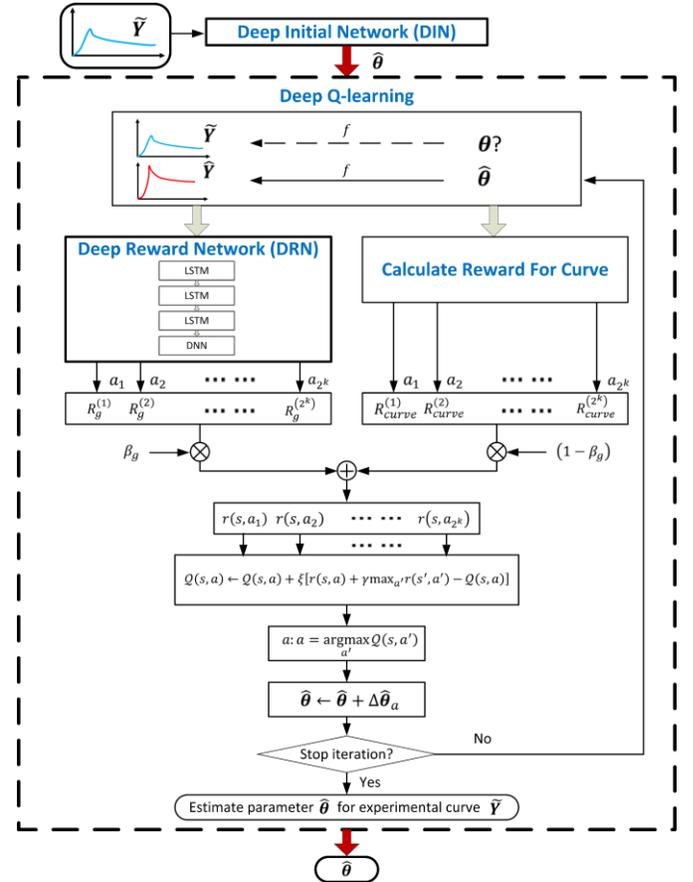

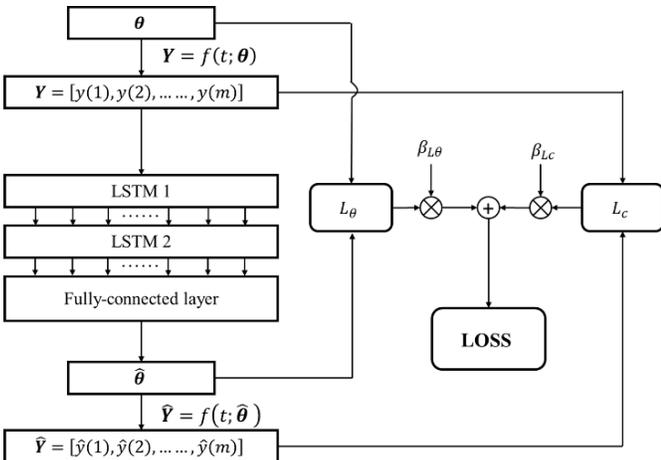

Fig. 3. Schematic illustration of the Deep Initial Network (DIN). The architecture of the DIN consists of LSTM layers and a fully connected layer. The last hidden layer applies is followed by a sigmoid nonlinearity; other hidden layers are followed by a rectifier nonlinearity.

Fig. 4. Schematic illustration of the Deep Q-learning of Model Parameters (DQMP) algorithm. DIN is used to predict initial guess of parameters, and DRN is used to predict global rewards $R_g$.

*B. KVFD model and its solution*

The KVFD model provides mathematically simple and experimentally flexible descriptions of viscoelastic properties of soft biological materials with physically interpretable model parameters [34]-[39]. The KVFD constitutive equation consists of a spring in parallel with a fractional-order dashpot element.

TABLE I
KVFD SOLUTIONS TO SPHERICAL INDENTATION TESTS

| Loading protocols | KVFD Solution | |
|---|---|---|
| Ramp-relaxation $P_{r,sphere}(t)$ | $\begin{cases} 4\sqrt{R}v^{\frac{3}{2}}E_0 t^{\frac{3}{2}}\left[\frac{2}{3} + \frac{\left(\frac{t}{\tau}\right)^{-\alpha}}{\Gamma(1-\alpha)}B\left(\frac{3}{2}, 1-\alpha\right)\right], & 0 < t \leq T_r \\ 4\sqrt{R}v^{\frac{3}{2}}E_0 T_r^{\frac{3}{2}}\left[\frac{2}{3} + \frac{\left(\frac{t}{T_r}\right)^{\frac{3}{2}}\left(\frac{t}{\tau}\right)^{-\alpha}}{\Gamma(1-\alpha)}B\left(\frac{T_r}{t}; \frac{3}{2}, 1-\alpha\right)\right], & t \geq T_r \end{cases}$ | (9-1) |
| Load-unload $P_{hys,sphere}(t)$ | $\begin{cases} 4\sqrt{R}v^{\frac{3}{2}}E_0 t^{\frac{3}{2}}\left[\frac{2}{3} + \frac{\left(\frac{t}{\tau}\right)^{-\alpha}}{\Gamma(1-\alpha)}B\left(\frac{3}{2}, 1-\alpha\right)\right], & 0 < t \leq T_r \\ 4\sqrt{R}v^{\frac{3}{2}}E_0 T_r^{\frac{3}{2}}\left[\frac{2}{3} + \frac{\left(\frac{t}{T_r}\right)^{\frac{3}{2}}\left(\frac{t}{\tau}\right)^{-\alpha}}{\Gamma(1-\alpha)}B\left(\frac{T_r}{t}; \frac{3}{2}, 1-\alpha\right)\right] - 4\sqrt{R}v^{\frac{3}{2}}E_0 \int_{T_r}^{t}\left[1 + \frac{((t-\xi)/\tau)^{-\alpha}}{\Gamma(1-\alpha)}\right](2T_r - \xi)^{\frac{1}{2}}d\xi, & T_r < t \leq 2T_r \end{cases}$ | (9-2) |
| Ramp-creep $h_r^{\frac{3}{2}}(t)$ | $\begin{cases} \frac{3}{8\sqrt{R}}\frac{kt}{E_0}\left[1 - \mathbf{E}_{\alpha,2}\left(-\left(\frac{t}{\tau}\right)^{\alpha}\right)\right], & 0 \leq t \leq T_r \\ \frac{3}{8\sqrt{R}}\frac{k}{E_0}\left[T_r + (t - T_r)\mathbf{E}_{\alpha,2}\left(-\left(\frac{t-T_r}{\tau}\right)^{\alpha}\right) - t\mathbf{E}_{\alpha,2}\left(-\left(\frac{t}{\tau}\right)^{\alpha}\right)\right], & t \geq T_r \end{cases}$ | (9-3) |

The constitutive equation relating stress $\sigma(t)$ to strain $\varepsilon(t)$ for the KVFD model is expressed as [39],

$$\sigma(t) = E_0 \varepsilon(t) + E_0 \tau^\alpha \frac{d^\alpha \varepsilon(t)}{dt^\alpha}, \quad (8)$$

where $E_0$ is an elastic modulus (Pa), $\tau(s)$ is a time constant, and $\alpha$ is a unitless real number between $(0,1)$ that defines the derivative order. The three model parameters, $E_0$, $\alpha$, and $\tau$, characterize respectively the elasticity, the fluidity, and the viscous time constant of the soft matter.

KVFD solutions to a variety of experimental protocols [39] collected using a spherical intender are the model functions applied during parameter estimation. These solutions are listed in Table I.

In (9-1)-(9-3), $T_r$ is the ramp time. $h_m$ and $P_m$ are the maximum displacement and force loadings. $v = h_m/T_r$, $k = P_m/T_r$. $B(x,y) = \int_0^1 t^{x-1}(1-t)^{y-1}dt$ (with $\mathrm{Re}(x) > 0$, $\mathrm{Re}(y) > 0$) is a complete beta function and $B(a;x,y) = \int_0^a t^{x-1}(1-t)^{y-1}dt$ for $a \in [0,1]$ is an incomplete beta function. In addition, $\mathbf{E}_{\alpha,\beta}(z) = \sum_{k=0}^{\infty}\frac{z^k}{\Gamma(\alpha k + \beta)}$ (with $\alpha > 0$, $\beta > 0$) is the Mittag-Leffler (M-L) function, for $\beta = 1$. It is written as the M-L function in one parameter, i.e. $\mathbf{E}_{\alpha,1}(z) \equiv \mathbf{E}_{\alpha}(z)$. The standard test setting is provided in Appendix B.

In the following, the DQMP algorithm is validated by fitting raw experimental data to the KVFD model equations.

## III. RESULT

The DQMP algorithm was applied to extract viscoelastic parameters $[E_0, \alpha, \tau]$ by fitting force mapping curves to KVFD solutions under different protocols (hereinafter referred to as KVFD equations, in which the fitting function $f$ is provided in (9-1)-(9-3) according to corresponding experimental protocols) and capability of DQMP algorithm was evaluated. The training data was generated by the KVFD modeling. The input of the curve sampled providing the time steps in the experiments.

All instances were set as follows: the experimental curves were sampled to $m$=250 points according to actual time-steps. To balance the importance of every input features, all parameters were normalized by the min-max normalization method [40]. For simulation, the KVFD parameter space range was $E_0 \in [10,100000]$, $\alpha \in [0.01,0.99]$ and $\tau \in [1,1000]$. The training data (1,000,000 curves) is randomly generated in parameters space. The detailed experimental settings is provided in Appendix C.

### A. Validation and evaluation of DQMP algorithm

#### 1) KVFD curve fitting

The simulation data was generated with the parameter $\boldsymbol{\theta} = [20000, 0.2, 50]$ by KVFD equations (9-1)-(9-3). The random noise was then added to the data. As shown in Fig. 5, the proposed method yielded a precise fit between the simulated data (blue circles) and the prediction curves (red lines) across all noise levels (rows) under all three loading protocols (columns), as the fitting is precise with goodness of fit $R^2 \geq 0.9912$, confirming that DQMP can fit the experimental data accurately. The estimated parameters listed in Table II are close to its global solutions as the relative errors of the estimated parameters are lower than 3% for $E_0, \alpha$ and $\tau$ respectively in all cases, indicating a great consistency of the estimated parameters and the true ones.

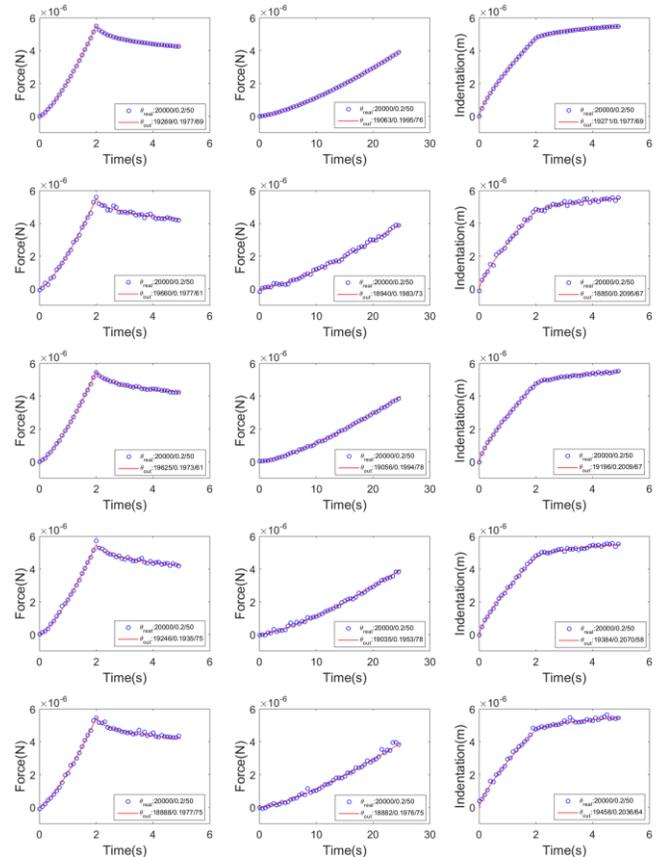

Fig. 5. KVFD fittings of the noisy simulation data generated with parameters $\boldsymbol{\theta} = [E_0, \alpha, \tau] = [20000, 0.2, 50]$. The curves in 1st-3rd columns are relaxation, load-unload, creep curves respectively, and curves in 1st-5th rows are the curve without noise, with Gaussian noise, Uniform noise, Rayleigh noise, Exponential noise respectively. The corresponding standard deviations of noise level are $10^{-7}$, $\frac{\sqrt{3}}{6} \times 10^{-7}$, $\frac{\sqrt{8-2\pi}}{2} \times 10^{-7}$ and $10^{-7}$, respectively. The simulated data are drawn with blue circles and the predicted fitting curves by DQMP are drawn with red lines.

TABLE II
RELATIVE ERRORS OF ESTIMATED PARAMETERS AND GOODNESS OF FIT, EACH BASED ON 500 RANDOMLY NOISY CURVES.

| Load protocol | Noise | Relative errors (%) of estimated parameters (mean ± std.) | | | $R^2$ (mean) |
|---|---|---|---|---|---|
| | | $\varepsilon_{E0}\%$ | $\varepsilon_\alpha\%$ | $\varepsilon_\tau\%$ | |
| Relaxation | Non | 0.6±0.4 | 0.3±0.2 | 1.6±1.2 | 0.9999 |
| | Gaussian | 0.5±0.4 | 0.4±0.3 | 1.5±1.4 | 0.9958 |
| | Uniform | 0.5±0.4 | 0.3±0.2 | 1.4±1.4 | 0.9996 |
| | Rayleigh | 0.5±0.4 | 0.4±0.3 | 1.5±1.3 | 0.9982 |
| | Exponential | 0.6±0.4 | 0.4±0.3 | 1.6±1.4 | 0.9955 |
| Load-unload | Non | 0.4±0.3 | 0.8±1.0 | 1.1±0.9 | 0.9999 |
| | Gaussian | 0.6±0.5 | 2.8±2.7 | 1.0±0.8 | 0.9929 |
| | Uniform | 0.5±0.4 | 2.0±2.0 | 1.1±0.8 | 0.9993 |
| | Rayleigh | 0.6±0.5 | 2.6±2.3 | 1.2±0.9 | 0.9969 |
| | Exponential | 0.7±0.6 | 3.0±2.6 | 1.2±0.9 | 0.9924 |
| Creep | Non | 0.9±0.7 | 0.5±1.1 | 2.7±3.0 | 0.9999 |
| | Gaussian | 0.9±0.7 | 0.8±1.3 | 2.6±2.6 | 0.9945 |
| | Uniform | 0.8±0.9 | 0.8±1.8 | 2.6±2.8 | 0.9993 |
| | Rayleigh | 0.9±0.7 | 0.6±1.2 | 2.6±2.4 | 0.9979 |
| | Exponential | 0.9±0.9 | 0.7±1.9 | 2.5±2.3 | 0.9948 |

*2) Evaluation of the estimated parameters by statistical analysis*

Ten thousand randomly curves (by adding Gaussian noise variance = $10^{-7}$ to the ideal curves to simulate experimental noisy data) were generated, and the parameters $[E_0, \alpha, \tau]$ were extracted by DQMP. The curve fitting was accurate with $R^2 \geq 0.952$. Pair-wise difference analysis and Pearson correlation analysis were conducted between the estimated parameters and the actual parameters. The results show that $E_0$ and $\alpha$ were close to their true values within 5% and 1.5% (p < 0.05). The estimated fluidity $\alpha$ (r = 0.999, p < 0.001) and elastic modulus $E_0$ (r=0.983, p < 0.001) were close to and highly correlated to their global solutions. To a lesser extent, the estimated viscosity $\tau$ (r = 0.866, p < 0.001) correlated to its global solutions, as $\tau$ is sensitive to noise.

*3) Evaluation of KVFD parameters imaging*

As shown in Fig. 6(a), the simulation image were generated with four sets of KVFD parameters [20000, 0.7, 800], [40000, 0.5, 600], [60000, 0.3, 400] and [80000, 0.1, 200] in the four 8 × 8 sub regions respectively. The corresponding four ideal curves generated and degraded by adding random Gaussian noise (variance = $10^{-6}$) to simulate the 256 experimental noisy curves. The initial guess of the parameters is provided by the proposed DIN.

The representative fitting of the noisy data by the proposed DQMP algorithm is shown in the first column of Fig. 6(b). All 256 noisy curves were fitted with $R^2 \geq 0.9643$. The estimated parameters were imaged as shown in the 2nd-4th column, from which we can see the extracted parameters were close to the ideal values and robust to Gaussian noise. The images of elastic modulus $E_0$ and fluidity $\alpha$ are almost uniform. The viscosity image of $\tau$ is with only slight noise fluctuation. All three imaged parameters can depict the true parameters well, which confirmed the accuracy and robustness of DQMP, indicating its potential of finding global parameters close to the true ones.

The results of DQMP (as displayed in Fig. 6(b)) is also compared with Q-learning (QL) and Least Square Method (LSM) optimization algorithms, respectively. As shown in Fig. 6(c), the Q-learning algorithm could also fit well as $R^2 \geq 0.9641$. However, with only curve fitting errors constraints, that is, there is no parameter constraints introduced, therefore, it equals to $R_g = 0$. The images of $E_0, \alpha, \tau$ are deviated to their ideal values, especially for $\tau$. As shown in Fig. 6(d), the LSM algorithm for extracting viscoelastic parameters show the largest deviation to their ideal values for $E_0$ and $\tau$.

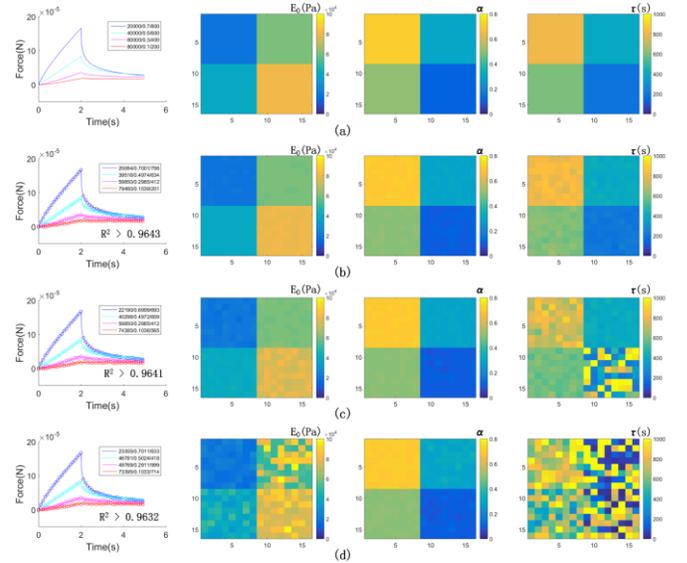

Fig. 6. Representative ramp relaxation curves generated with KVFD parameters [20000, 0.7, 800], [40000, 0.5, 600], [60000, 0.3, 400] and [80000, 0.1, 200] and the fit of noisy curves by DQMP, QL and LSM algorithms (top to bottom), respectively. The corresponding viscoelastic parameters $[E_0, \alpha, \tau]$ in 16 × 16 matrix (left to right, respectively) are (a) simulation parameters image, (b) imaged parameters by DQMP, (c) by QL and (d) by LSM algorithms.

TABLE III
RELATIVE ERRORS FROM DIFFERENT FITTING METHODS

| Algorithm | Relative errors (%) of estimated parameters (mean ± std.) | | |
|---|---|---|---|
| | $\varepsilon_{E_0}\%$ | $\varepsilon_{\alpha}\%$ | $\varepsilon_{\tau}\%$ |
| DQMP | -0.08 ± 0.49 | -0.05 ± 0.27 | 0.88 ± 1.71 |
| QL | -4.10 ± 2.34 | -0.13 ± 0.35 | 8.10 ± 23.94 |
| LSM | 1.98 ± 9.97 | 0.14 ± 1.05 | 10.46 ± 33.93 |

The errors image $\Delta x = \hat{x} - x$ are analyzed. As listed in Table III, the proposed DQMP performs best with the least mean errors (<2%) compared to QL and LSM in all $E_0$, $\alpha$ and $\tau$ parameters. The parameters are consistent as they are all in a close agreement to global solutions. Q-learning is with mediate precision where $\tau$ is not consistent, and LSM show inconsistency in the three imaged parameters far from to their global solutions, due to trapping into different local minima and making the images unreliable.

### B. Imaging applications

The proposed DQMP was further validated by viscoelastic imaging on a variety of soft matters under different loading protocols. The viscoelastic parameters estimated using the proposed DQMP for all tested points are imaged.

As shown in Fig. 7, Gelatin-Cream samples were tested under the relaxation loading by the nanoindentation with spherical indenter [39]. For each sample, a total of 20 ×20 (for imaging purposes) array of testing points were acquired. The curve fittings by the proposed DQMP are accurate with $R^2 \geq 0.9418$. The montage images of the two samples show obvious

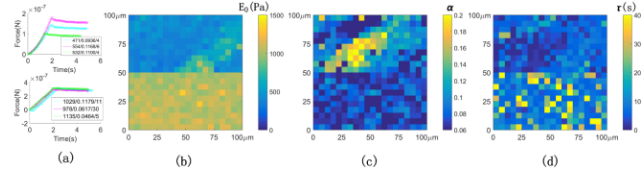

Fig. 7. Imaging of Gelatin-Cream samples. (a) The representative relaxation curves tested from Gel5Cream0 and Gel5Cream25 samples fitted by the proposed DQMP. The experimental data are drawn with circles and the predicted curves are drawn with lines. Viscoelastic parameters $[E_0, \alpha, \tau]$ are imaged shown in 20×20 pixel regions (100μm × 100μm) in (b)-(d), where Gel5Cream25 sample (upper half) and Gel5Cream0 (lower half) are differentiable.

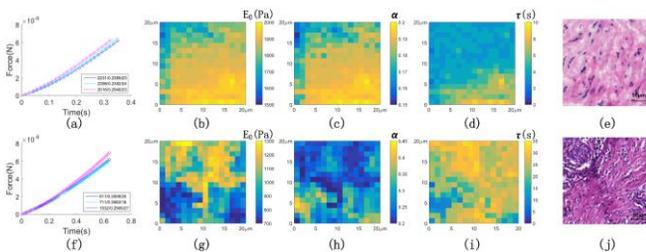

Fig. 8. Imaging of breast-tissue biopsy samples. (a) The representative approaching curves from the load-unload testing for benign tissue fitted by the proposed DQMP. The experimental data are drawn with circles and the predicted curves are drawn with lines. (b)-(d) are 20μm × 20μm images of viscoelastic quantities $[E_0, \alpha, \tau]$ (left to right, respectively) for ductal epithelial hyperplasia tissue. (e) is the histological image for a much larger region that includes the parametric images. A dark square (20μm × 20μm, 256 pixels) represents the size of the viscoelasticity map in HE staining image. (f)-(j) are for invasive carcinoma tissue. The total length of scale bars in the histology image is 50μm.

difference in the montage interface.

As shown in Fig. 8, the AFM force map of the punctured breast tissue was acquired under load-unload testing with a spherical indenter (Ethics number: XJTU1AF2017LSK-46). All curves fitting were accurate with $R^2 \geq 0.9708$. Images of viscoelastic parameters $[E_0, \alpha, \tau]$ for benign (upper) and malignant (lower) tissue samples in $16 \times 16$ pixel regions (20μm × 20μm) are provided in the second to fourth column. There were significant differences found between benign tissue versus malignant tissue based on the elastic modulus $E_0$ (p < 0.001), the fluidity $\alpha$ (p < 0.001) and the viscous time constant $\tau$ (p < 0.001) .It can be seen that benign tissue is regionally homogenous in its viscoelastic properties, while malignant tissue is more heterogeneous.

Furthermore, ultrasound viscoelastic imaging of breast tissue is validated. The data collected under the condition of creep loading was fitted to ramp creep solutions with a plate indenter as follows [39],

$$\varepsilon_r(t) = \begin{cases} \frac{\sigma_0}{E_0} \frac{t}{T_r} \left[1 - \mathbf{E}_{\alpha,2}\left(-\left(\frac{t}{\tau}\right)^\alpha\right)\right], & 0 \leq t \leq T_r \\ \frac{\sigma_0}{E_0} \frac{1}{T_r} \left[T_r + (t-T_r)\mathbf{E}_{\alpha,2}\left(-\left(\frac{t-T_r}{\tau}\right)^\alpha\right) - t\mathbf{E}_{\alpha,2}\left(-\left(\frac{t}{\tau}\right)^\alpha\right)\right], & t \geq T_r \end{cases}, (10)$$

where $\sigma_0$ is the maximum loading stress, and $\varepsilon_r$ is the response strain.

All curves fitting were accurate with $R^2 \geq 0.9831$. Fig. 9(b)-(d) and Fig. 9(f)-(h) are in-vivo images of the viscoelastic parameters $[E_0, \alpha, \tau]$ displayed as color overlays onto in-vivo ultrasonic strain images of benign and malignant breast tissues, respectively. The results indicate malignant tissue has higher elastic modulus and fluidity whereas lower viscosity as compared to its benign counterpart, which provides valuable image information for diagnosis of disease.

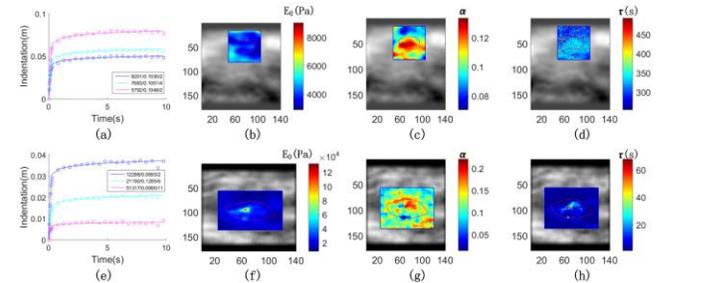

Fig. 9. Ultrasound viscoelastic imaging of breast tissue and creep curves fitting result. The representative creep curves taken from benign tissue (a) and malignant tissue (e) fitted by the proposed DQMP, the experimental data are drawn with circles and the predicted curves are drawn with lines. The viscoelastic parameters $[E_0, \alpha, \tau]$ are extracted by the DQMP algorithm by fitting to a KVFD creep model. (b)-(d) are viscoelastic parameter images superimposed onto ultrasonic strain images for benign lesions. (f)-(h) are results for malignant breast tissues.

## IV. DISCUSSION

We demonstrated the efficacy of our proposed method via simulation and viscoelasticity imaging.

First, DQMP is assessed and validated by simulation data under different protocols by KVFD modeling. The proposed DQMP method could find global model parameters promisingly. Statistical analysis on 10,000 curves showed that all curves were fit accurately with $R^2 \geq 0.952$, and $E_0$ and $\alpha$ were close to their true values within 5% and 1.5% (p < 0.05) in relative errors. Moreover, DQMP is also robust to different types of noise. Among Gaussian, Uniform, Rayleigh, and Exponential noise, DQMP performs best for curves with Uniform noise.

Second, simulation imaging demonstrated the consistency and reliability of the parameters image. As compared with QL and LSM, the parameters imaged by DQMP were closer to the ground truth image (<2% in relative error). The proposed DQMP are with the least errors for imaging parameters, showing the powerful global searching ability of DQMP.

Third, DQMP is applied to viscoelastic imaging on a variety of soft matters under different loading protocols. The results indicate a great potential of DQMP for imaging of physical parameters in diagnosis purpose (i.e., benign vs. malignant). Imaging on breast tissue samples further validates the robustness, applicability, and efficacy of the proposed DQMP in that the increased elastic modulus and fluidity for malignant tissues compared to benign tissue provides with valuable information for disease diagnosis.

## V. Conclusion

This is the first report to formulate model parameter optimization task to be a state-action decision making in the parameter space. We leveraged Q-learning learning integrated with deep learning to build a model designed to learn global reward values from both visible state (curve fitting) and hidden state (parameters fitting), and proposed DQMP scheme for global parameters optimization on complex nonconvex function. Through DQMP, parameters searching resembles an action selection ($k$-D move) from $2^k$ configurations in the $k$-dimensional parameters space. The proposed DQMP combines the prior knowledge through DRN. Appropriate decision is made by maximizing Q-value, which combines the current reward and future reward functions from both visible states and hidden states, so as to update parameters toward global solution iteratively.

In summary, the novelty of the work is as follows:
1) Model parameter optimization problem is converted to be state-action decision making in the parameters space.
2) To guide global searching, a deep reward network is proposed to learn the global reward from both hidden state and visible state.
3) A novel DQMP method considers not only current reward function but also future reward function, so as to lead global searching iteratively by maximizing Q-value in the parameter space.

The proposed DQMP is demonstrated to be capable of finding global optional model parameters and show potential for imaging of physical parameters in many applications.

While the proposed DQMP was tested in KVFD model in this study, the proposed frameworks can be generalized to global optimization for other complex nonconvex functions and physical parameters imaging. In particular, in the field of ultrasonic mechanical imaging, the DQMP is expected to obtain reliable viscoelastic imaging. DQMP could also accurately find global model parameters with high accuracy and consistency, which is crucial for the development of imaging algorithm and equipment.

## Appendix

### A. DQMP algorithm

The DQMP algorithm is described as follows.

Firstly, initial guess of parameters $\hat{\boldsymbol{\theta}}$ for experimental curve $\widetilde{\boldsymbol{Y}}$, also referring to desired curve, is predicted by Deep Initial Network (DIN). Given the immediate curve (determined by $f(t;\hat{\boldsymbol{\theta}})$) and the desired curve, global reward value $R_g$ was predicted by DRN. In each DQMP iteration, $r(s,a)$ and $\max_{a'} Q(s',a')$ for current state were calculated by combining predicted global reward $R_g$ and immediate curve fitting reward $R_{curve}$. Q-value is updated according to (2), and action of parameters updating is conducted by maximizing Q-value policy. The algorithmic flow of DQMP is provided in Algorithm 1.

### B. Standard indentation test

As shown in Fig. A.1, the responses of samples (lower row of curves) to a force or displacement input over time (upper row of curves) is recorded by the indenter for stress-relaxation, load-unload or creep-recovery experiments. These are the raw data used to estimate stress-strain behavior and ultimately the viscoelastic properties. $T_r$ is the ramp time. $v = h_m/T_r$, $k = P_m/T_r$.

### C. Experimental settings of different protocols performed in the experiments

The detailed experimental settings of different protocols performed in the experiments are provided in Table A.I.

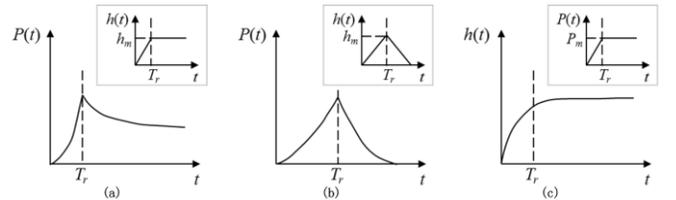

Fig. A.1. Schematic illustration of (a) ramp-hold relaxation for ramp-on time $T_r$, (b) load-unload and (c) ramp-hold creep protocols. The excitation signals are shown in small gray boxes above the response signals, where $t$ is time, $h$ is indentation depth and $P$ is force.

**Algorithm 1:** Algorithmic flow of DQMP. $\widehat{\boldsymbol{\theta}} = \text{DQMP}(\widetilde{\boldsymbol{Y}})$

**Input**:
  $\widetilde{\boldsymbol{Y}}$ – Experimental data; $\widetilde{\boldsymbol{Y}} \approx \boldsymbol{Y} = f(t; \boldsymbol{\theta})$; $f$ is the complex, nonconvex mathematical modeling function.
  $\boldsymbol{\theta}$ is global optimal parameters

**Output**:
  $\widehat{\boldsymbol{\theta}}$ – Estimated global optimal parameters approaching the global optimal solution $\boldsymbol{\theta}$

1: $j = 1$
2: Initialize Q-table
3: $\widehat{\boldsymbol{\theta}}^{(1)} = \text{DIN}(\widetilde{\boldsymbol{Y}})$   // DIN is Deep Initial Network, see section 2.1.3.
4: **while** (~convergence)
5:   $\widehat{\boldsymbol{Y}}^{(j)} = f(t; \widehat{\boldsymbol{\theta}}^{(j)})$
6:   $\boldsymbol{R}_g = \text{DRN}(\widehat{\boldsymbol{Y}}^{(j)} - \widetilde{\boldsymbol{Y}})$   // DRN is Deep Reward Network, see section 2.1.2, $\boldsymbol{R}_g$ is a vector
7:   **for** all actions $a$: $(a = a_1, \dots, a_{2^k})$
8:     $R_{curve,a} \leftarrow g\left(\mathbf{E}\left(|\widehat{\boldsymbol{Y}}_a^{(j)} - \widetilde{\boldsymbol{Y}}|\right)\right)$   // $\widehat{\boldsymbol{Y}}_a^{(j)}$ is the estimation of $\boldsymbol{Y}$ when selecting action $a$
9:     $r(s,a) \leftarrow \beta_g R_{g,a} + (1-\beta_g) R_{curve,a}$   // $R_{g,a}$ is the global rewards corresponding to action $a$
10:    call $\text{DRN}(\widehat{\boldsymbol{Y}}_a^{(j)} - \widetilde{\boldsymbol{Y}})$ to predict $\boldsymbol{R}_{g,a'}$
11:    **for** all actions $a'$: $(a' = a_1', \dots, a_{2^k}')$
12:      calculate $r(s',a')$ with $R_{g,a'}$
13:    **end for**
14:    $Q(s,a) \leftarrow Q(s,a) + \xi[r(s,a) + \gamma \max_{a'} r(s',a') - Q(s,a)]$
15:  **end for**
16:  choose action $a$: $a \leftarrow \arg\max_a Q(s,a)$
17:  $\widehat{\boldsymbol{\theta}}^{(j+1)} \leftarrow \widehat{\boldsymbol{\theta}}^{(j)} + \Delta \widehat{\boldsymbol{\theta}}_a$
18:  $j \leftarrow j + 1$
19: **end while**
20: **return** $\widehat{\boldsymbol{\theta}} = \widehat{\boldsymbol{\theta}}^{(j)}$

TABLE A.I
DETAILED EXPERIMENTAL SETTINGS OF DIFFERENT PROTOCOLS PERFORMED IN THE EXPERIMENTS

| Experiment category | Experimental settings | | | | | |
|---|---|---|---|---|---|---|
| | Protocols | Probe type | Probe shape | Loading range | Ramp time $T_r$ | Hold time $T_{hold}$ |
| Simulation data | Relaxation | Spherical | Radius (8.5μm) | Depth (5μm) | 2s | 3s |
| | Load-unload | Spherical | Radius (8.5μm) | Depth (5μm) | 25s | / |
| | Creep | Spherical | Radius (8.5μm) | Stress (5μN) | 2s | 3s |
| Nanoindentation data | Relaxation | Spherical | Radius (8.5μm) | Depth (5μm) | 2s | 3s |
| | Load-unload | Spherical | Radius (5μm) | Depth (3.6μm) | 1.2s | / |
| Ultrasonic data | Creep | Plate | Area (24cm$^2$) | Stress (833.3Pa) | 0.25s | 9.75s |